\documentclass{article}
\usepackage[T1]{fontenc}
\usepackage[utf8]{inputenc}
\usepackage{microtype}
\usepackage{graphicx}
\usepackage{booktabs,tabularx,array}
\usepackage{multirow}
\usepackage{xurl}
\usepackage{hyperref}

\usepackage[accepted]{icml2025}
\makeatletter
\renewcommand{\Notice@String}{Preprint. Work in progress.}
\makeatother
\usepackage{amsmath,amssymb,mathtools,amsthm}
\usepackage[capitalize,noabbrev]{cleveref}
\graphicspath{{figures/}}
\icmltitlerunning{Think with Skills: Pseudocode Planning Before Execution}

\begin{document}
\twocolumn[
\icmltitle{Think with Skills: Pseudocode Planning Before Execution for Advanced Reasoners}
\icmlsetsymbol{equal}{*}
\begin{icmlauthorlist}
\icmlauthor{Boshen Zhou}{equal,honor}
\icmlauthor{Boxuan Yue}{equal,honor}
\icmlauthor{Xiang Li}{honor}
\icmlauthor{Shanchao Yang}{honor}
\end{icmlauthorlist}
\icmlaffiliation{honor}{Honor Device Co., Ltd., China. Emails: \nolinkurl{quanlang2017@outlook.com}, \nolinkurl{yueboxuan@honor.com}, \nolinkurl{lixiang27@honor.com}, and \nolinkurl{yangshanchao@honor.com}}
\icmlcorrespondingauthor{Boxuan Yue}{yueboxuan@honor.com}
\vskip 0.3in
]
\printAffiliationsAndNotice{\icmlEqualContribution}

\begin{abstract}

Language model agents often struggle with complex tasks even when equipped with expert-written skills. Such skills encode declarative knowledge, whereas successful execution requires procedural knowledge tailored to the task at hand. Natural language skills may leave conditions and dependencies implicit, and tasks involving multiple skills further require agents to coordinate intermediate results and execution order. We introduce Think with Skills, a method that trains reasoning models to turn relevant skill descriptions into task-specific pseudocode plans before execution. The pseudocode explicitly represents operations, conditions, intermediate results, and dependencies, providing a structured bridge from skill descriptions to agent actions. We construct training data that pairs tasks and skills with pseudocode plans and execution trajectories, use supervised fine tuning to initialize this behavior, and further optimize end-to-end task performance with Group Relative Policy Optimization. On SkillsBench, Think with Skills raises task success from 35.0\% to 62.0\%. These results show that proceduralizing expert knowledge before execution substantially improves skill use.

\end{abstract}
\section{Introduction}
\label{sec:1}

\begin{figure*}[t]
\centering
\includegraphics[width=\linewidth,height=0.65\textheight,keepaspectratio]{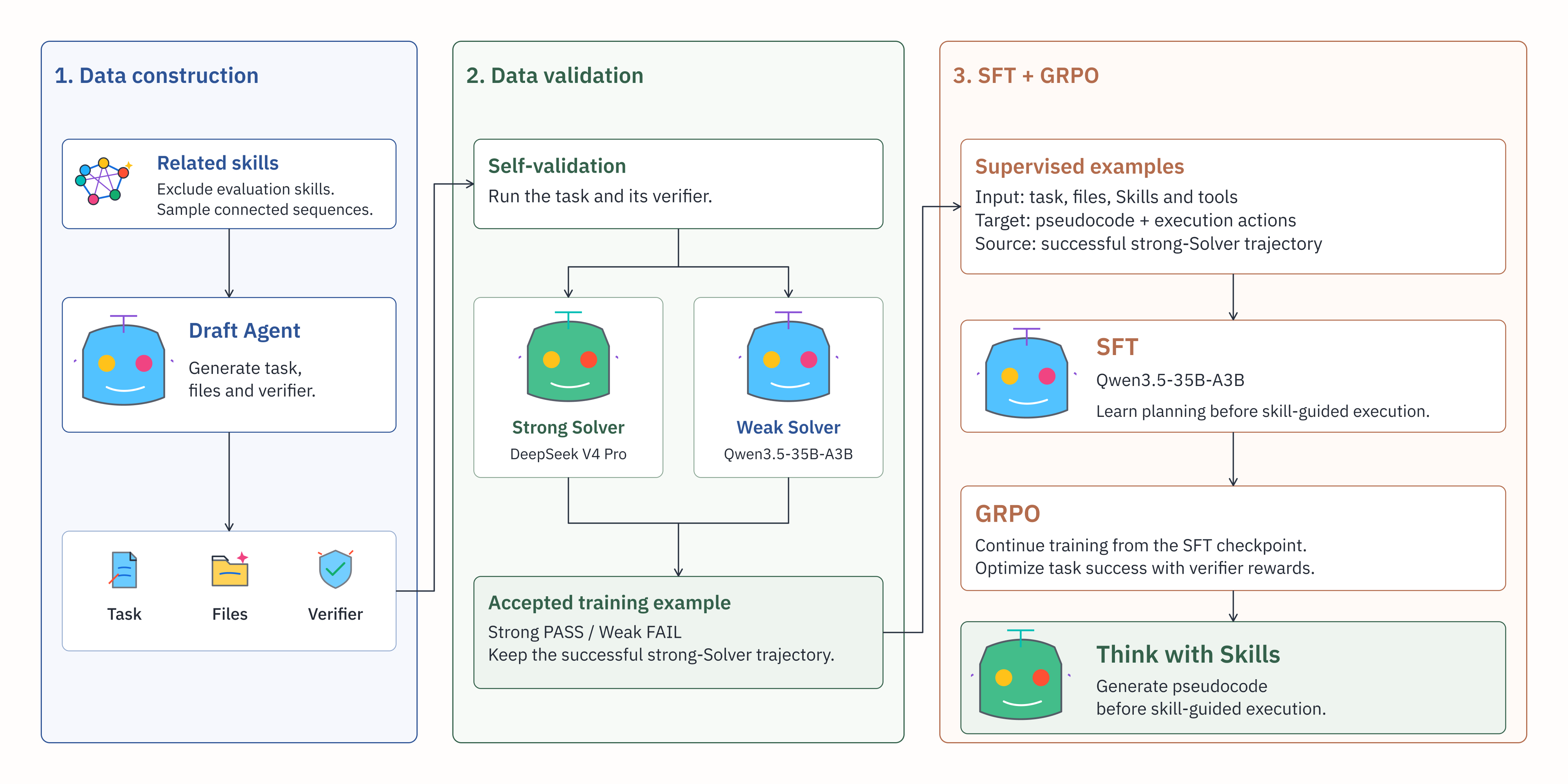}
\caption{Overview of the Think with Skills training pipeline. A Draft Agent constructs and validates tasks from related skills. Strong and weak Solvers then attempt each task. Tasks solved by the strong Solver but not the weak Solver are retained, and the corresponding trajectories provide pseudocode and execution supervision for SFT. GRPO continues training from the SFT checkpoint with verifier rewards.}
\label{fig:1}
\end{figure*}

A medical student may memorize the symptoms of many diseases and detailed descriptions of surgical procedures, yet still require extensive clinical training before becoming a qualified physician. Memorization gives the student declarative knowledge about medicine, while clinical practice develops the procedural knowledge needed to apply it to real patients. Cognitive theories distinguish these two forms of knowledge: declarative knowledge represents facts and explicit instructions, whereas procedural knowledge determines how that information guides action \citep{anderson1982acquisition}. Completing a complex task requires both knowing what is relevant and knowing how to apply that knowledge in the current situation.

Large language models are increasingly used as agents that reason, invoke tools, observe their results, and decide what to do next \citep{yao2023react}.  To provide agents with expertise for specialized tasks, recent systems augment them with Agent Skills. A skill packages domain knowledge, instructions, and tool usage conventions written by human experts, allowing an agent to receive reusable task guidance without changing its parameters \citep{li2026skillsbench}. Curated skills can substantially improve task success, but their benefits vary widely across tasks, and access to a skill does not guarantee correct execution. From the agent's perspective, the contents of a skill remain natural language descriptions. They make expert knowledge available, but the agent must still learn how to apply that knowledge in the current task.

Turning a skill description into action is difficult for several reasons. First, guidance written by human experts may describe execution logic incompletely and leave conditions, intermediate states, and dependencies implicit. Second, a skill usually describes a general procedure, while the agent must adapt it to the current objective, available tools, files, observations, and intermediate results. Third, a complex task may require several skills whose procedures were written independently. The agent must determine when each skill applies, connect the output of one skill to the input of another, and establish a coherent execution order. Studies of natural language described algorithms similarly suggest that understanding an instruction and reliably carrying out its control flow are separate challenges for language models \citep{zheng2024executing}. We call the gap between receiving expert guidance and turning it into an executable procedure for the current task the skill proceduralization gap.

Research on intermediate reasoning suggests a possible path forward. Chain of thought prompting allows models to express reasoning steps before producing an answer \citep{wei2022chain}, while least to most prompting and Plan and Solve separate problem decomposition from solution generation \citep{zhou2023least,wang2023plan}. Program aided methods use code as an intermediate representation for numerical and symbolic reasoning \citep{gao2023pal,chen2023program}. Think and Execute extracts logic shared across instances of an algorithmic task, represents it as task level pseudocode, and asks a language model to simulate its execution \citep{chae2024language}. These methods show that both prior thinking and structured intermediate representations can improve reasoning. However, they do not teach agents to convert one or more external skills into an instance specific procedure grounded in the current environment.

We introduce Think with Skills, a training approach that teaches reasoning models to perform this conversion before execution. Given a task and its relevant skills, the model first generates a task specific pseudocode plan in its thinking process. The plan makes operations, applicable conditions, execution order, intermediate results, and dependencies across skills explicit. The agent then follows this plan while interacting with the environment. In this way, pseudocode serves as a structured intermediate representation between natural language skill descriptions and concrete agent actions, allowing general expert guidance to be organized around the requirements of the current task.

Existing rationale supervision primarily teaches models to derive answers from questions \citep{zelikman2022star,hsieh2023distilling}. To supervise skill proceduralization, we construct training examples that pair tasks and relevant skills with pseudocode plans and their corresponding execution trajectories. We first use these examples for supervised fine tuning, teaching the model to generate a plan from the provided skills and carry out the task according to that plan. We then apply Group Relative Policy Optimization (GRPO) \citep{shao2024deepseekmath}, using executable verification feedback to optimize whether the complete planning-and-execution process leads to task success. This two-stage procedure combines explicit trajectory supervision with outcome-based optimization, as summarized in Figure~\ref{fig:1}.

We apply Think with Skills to Qwen3.5-35B-A3B and evaluate the resulting model on SkillsBench. Our two-stage training procedure increases the task success rate from 35.0\% to 62.0\%, an improvement of 27.0 percentage points. Our contributions are threefold. First, we formulate the skill proceduralization gap and introduce pseudocode planning as an intermediate step between reading skills and executing them. Second, we develop a data construction method and training procedure that teach models to convert expert skills into task specific plans and execution trajectories. Third, we demonstrate that learning this behavior substantially improves skill use on SkillsBench, showing that expert guidance becomes more effective when agents first organize it into an explicit procedure for the current task.

\section{Related Work}
\label{sec:2}

\subsection{Agent Skills}
\label{sec:2.1}

Agent skills are reusable artifacts that encode knowledge and procedures for recurring tasks as natural language instructions, tool-use conventions, or executable programs. They allow agents to reuse prior experience without deriving the complete procedure for every new instance. SkillsBench shows that human-curated skills improve the average pass rate by 16.2 percentage points across diverse tasks and agent configurations \citep{li2026skillsbench}.

\subsubsection*{Skill Construction and Refinement}

Prior work has studied both skill acquisition and skill refinement. Demonstration- and experience-based approaches abstract reusable procedures from task demonstrations or accumulated interactions \citep{liu2024skillact,zhao2024expel}. Other methods improve these procedures through reinforcement learning, execution-trace-based revision, or the compilation of heterogeneous external knowledge into structured skill representations \citep{wang2025sage,liu2026skillrevise,pan2026anything2skill}. This line of work broadens the sources of reusable skills and improves the quality of their representations. How a general procedure should be grounded in the task-specific objective, input files, available tools, and evolving observations is still left to the executor.

\subsubsection*{Skill Retrieval and Composition}

Voyager represents each acquired skill as executable code, indexes it by an embedding of an automatically generated description, and retrieves the five skills most similar to a query formed from a task-solving suggestion and the current environment feedback \citep{wang2024voyager}. SRA studies retrieval from a much larger external corpus. Sparse or dense retrievers produce a candidate set, an optional LLM reranker refines its ordering, and the agent either receives the top-ranked skill directly or selects which full instructions to load from a metadata-only catalog \citep{su2026sra}. SkillComposer goes beyond independently ranking candidates by encoding the task, environment, and compact skill metadata, then autoregressively decoding a variable-length sequence of skill identifiers whose membership, cardinality, and order are predicted jointly \citep{zhao2026skillcomposer}. These methods improve access to relevant skills and provide a coarse execution order, but they do not instantiate the selected instructions as a task-specific procedure that binds each step to current inputs and tools, specifies the flow of intermediate results, and adapts its control flow to new observations.

\subsubsection*{Skill Execution Control}

Execution-oriented methods add explicit constraints to the supplied procedure. Grounded Skill-Following uses a contract runtime to define required phases, legal actions, state transitions, and accepted termination, making procedural compliance verifiable during execution \citep{shen2026grounded}. Such verification depends on a formal contract and a state-tracking runtime supplied alongside the natural language skill. Ordinary expert-written skills provide neither structure by default, leaving task-specific conditions, intermediate states, and dependencies implicit.

\subsection{Planning and Programmatic Reasoning}
\label{sec:2.2}

Planning methods organize agent behavior either by interleaving reasoning with actions and feedback-based replanning, or by separating a high-level planner from a low-level executor \citep{yao2023react,liu2024rafa,erdogan2025planandact}. Recent post-training approaches further learn planning from synthesized or distilled trajectories and favor examples whose plans align with their execution \citep{parmar2025plantuning,liu2026cope}. Complementary work studies how such reasoning procedures should be represented. Program-aided methods translate natural language problems into executable programs that are run by an external interpreter \citep{gao2023pal,chen2023program}, whereas pseudocode-based methods express reusable task logic as a structured intermediate representation for the model to simulate \citep{chae2024language,zheng2024executing}. Across both lines of work, plans and programs are generally derived from task descriptions, worked examples, or interaction trajectories. They do not address how an external expert skill should be converted into pseudocode tailored to the current objective, inputs, and tools before the agent begins acting.

\section{Method}
\label{sec:3}

\begin{figure*}[t]
\centering
\includegraphics[width=\linewidth,height=0.65\textheight,keepaspectratio]{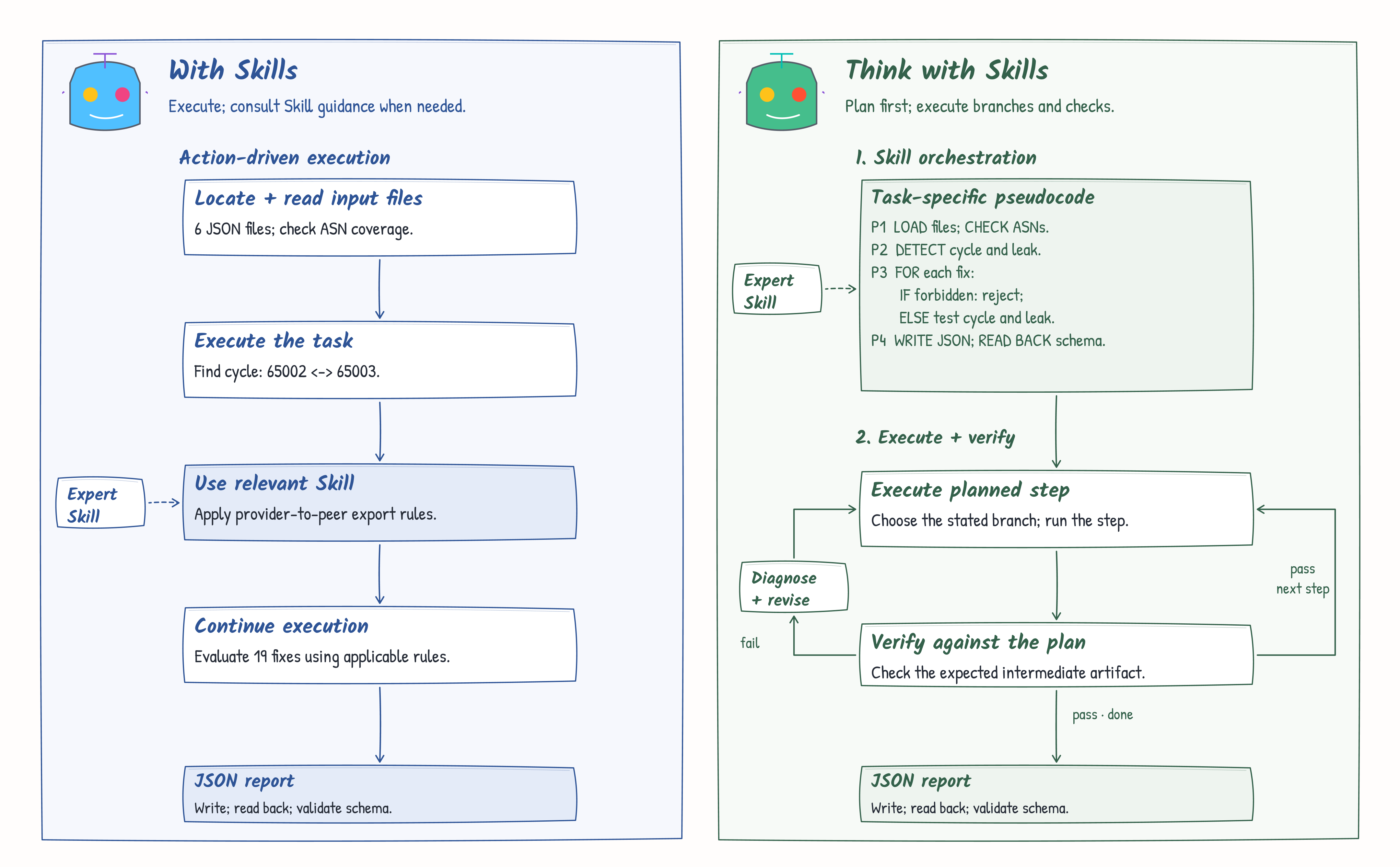}
\caption{Illustrative comparison of direct skill use and Think with Skills on a routing-diagnosis task with the same input files and expert skill. Direct skill use consults relevant guidance during execution. Think with Skills first expresses the skill as task-specific pseudocode, then follows its branches and checks while executing. Failed checks trigger diagnosis and revision of the affected step. Passed checks allow execution to continue or finish.}
\label{fig:2}
\end{figure*}

Think with Skills trains an agent to proceduralize expert-written skills before interacting with tools. Given a task, its input files, the available tools, and one or more relevant skills, the agent first transforms the skills into task-specific pseudocode and then executes the resulting procedure. Our method has three components. First, skill proceduralization converts general natural-language guidance into a structured procedure grounded in the current task. Second, a skill-grounded data pipeline constructs paired pseudocode plans and execution trajectories. Third, a two-stage training procedure uses supervised fine-tuning to initialize planning and execution, followed by GRPO to optimize end-to-end task success. Figure~\ref{fig:1} summarizes the overall workflow, Figure~\ref{fig:2} illustrates proceduralized execution through a routing-diagnosis example, and Figure~\ref{fig:3} expands the data-construction and validation stages.

\subsection{Pre-Execution Skill Proceduralization}
\label{sec:3.1}

As shown in Figure~\ref{fig:2}, we separate skill use into task-specific pseudocode construction and plan-aligned execution. Expert-written skills are usually expressed as natural-language guidance. They may describe useful operations without specifying how those operations should be instantiated for the current files, tools, and intermediate states. Before performing any skill-guided work, Think with Skills requires the model to rewrite the relevant guidance as task-specific pseudocode. This pseudocode is not a summary of the skill. It is a prospective execution specification that states how the skill will be applied to the current task.

\textbf{Pseudocode specification.} Each plan must contain five elements. First, it decomposes the intended workflow into explicitly numbered steps, with each step describing a concrete operation rather than a general objective. Second, it identifies the inputs required by each step and the output or artifact that the step is expected to produce. Third, it represents control flow explicitly: if/else branches specify how execution should respond to alternative observations, and loops specify when an operation should be repeated. Fourth, it defines the prerequisites and dependencies of later steps, including which intermediate result must be available before execution can continue. Fifth, it states success, failure, and verification conditions for the produced artifacts. The resulting plan therefore exposes operation order, data flow, control flow, and completion criteria that may remain implicit in the original skill description.

\textbf{Plan-aligned execution.} The pseudocode remains available throughout execution and is used to constrain each subsequent action. Before acting, the model identifies the numbered step and control-flow branch currently being executed, checks whether its prerequisites have been satisfied, and specifies the expected artifact. After acting, it verifies the observed result against the condition written in the plan. Execution advances only after the current output has been validated. When validation fails, the model attributes the failure to the relevant part of the procedure, such as an incorrect branch, a missing prerequisite, an invalid assumption, or an execution error, and revises the affected step instead of repeating the same action without diagnosis.

This representation has three properties. It is \emph{skill-grounded} because its operations are derived from the supplied expert guidance. It is \emph{instance-specific} because abstract instructions are bound to the current task, files, tools, and observations. It is \emph{verifiable} because intermediate artifacts and completion conditions are stated before execution. For tasks involving multiple skills, named intermediate artifacts make the output of one procedure available as the input to another and clarify their execution dependencies. The pseudocode is not passed to an external interpreter. It functions as an explicit reasoning scaffold that the model follows and updates while interacting with the environment.

\subsection{Skill-Grounded Data Construction}
\label{sec:3.2}

\begin{figure*}[t]
\centering
\includegraphics[width=\linewidth,height=0.65\textheight,keepaspectratio]{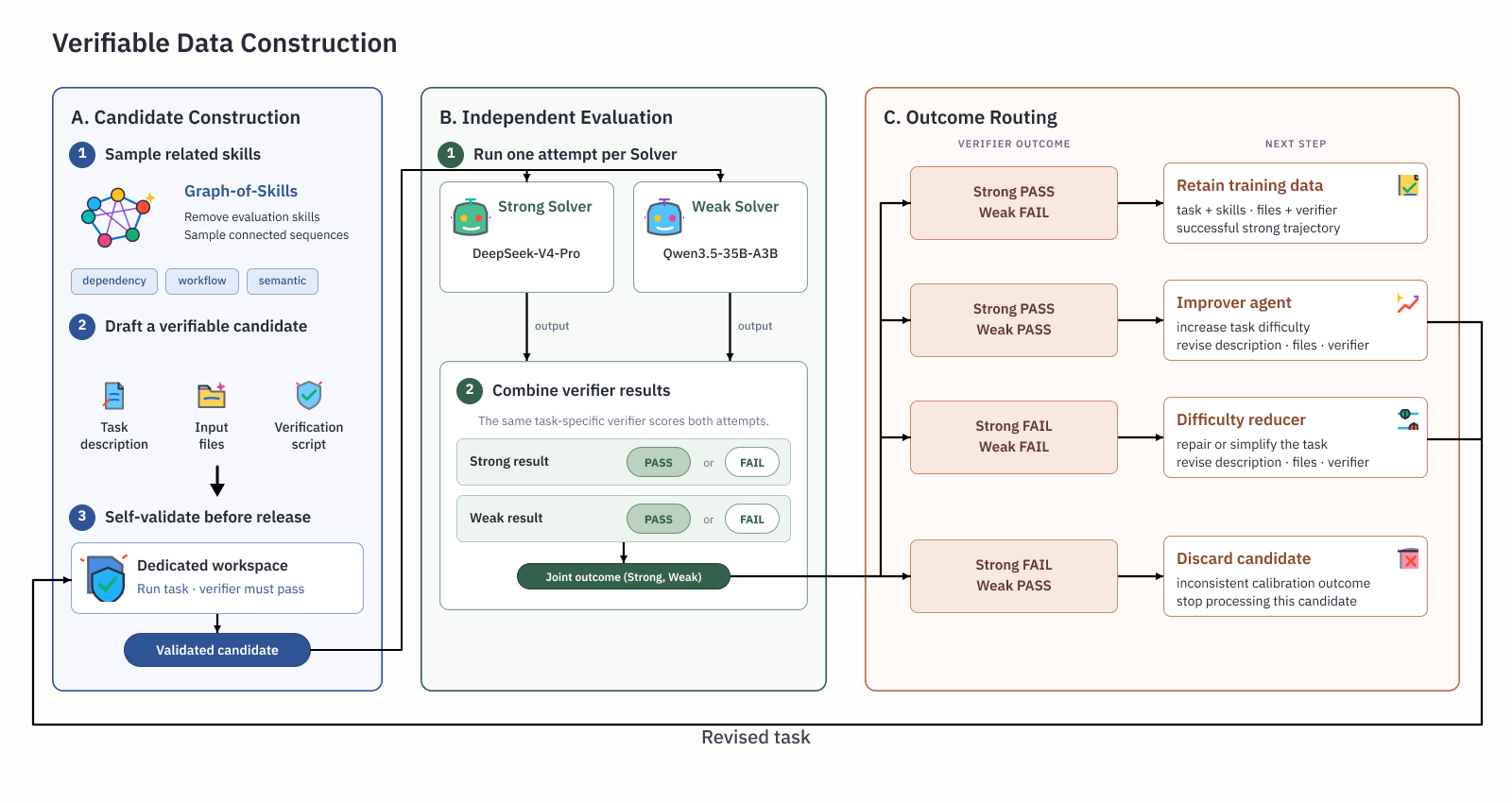}
\caption{Verifiable and difficulty-calibrated data construction. (A) A Draft Agent constructs a task, input files, and verifier from related skills and self-validates the candidate. (B) The strong and weak Solvers independently attempt the same candidate once, and the task-specific verifier scores both outputs. (C) Their joint outcome determines whether to retain the candidate, increase its difficulty, repair or simplify it, or discard it. Revised tasks return to self-validation before another independent evaluation by both Solvers.}
\label{fig:3}
\end{figure*}

We construct tasks that are automatically verifiable, solvable by a stronger model, and challenging for the target model. Figure~\ref{fig:3} organizes the pipeline into candidate construction, independent Solver evaluation, and outcome routing. We begin with Graph-of-Skills (GoS), a typed directed graph in which skills are connected by dependency, workflow, semantic, and alternative relations \citep{liu2026graphskills}. After removing skills reserved for evaluation, we sample connected skill sequences from the remaining graph. These sequences provide a structural basis for generating tasks that require one or more related skills rather than arbitrary skill combinations.

\textbf{Candidate construction and self-validation.} The Draft Agent receives the sampled skills and freely constructs a task description, its input files, and a verification script without relying on a predefined task template. The verifier provides an executable criterion for task completion. The Draft Agent must run the task once in its own workspace and pass the verifier before the candidate proceeds to model-based validation. This step filters out tasks with inconsistent instructions, malformed inputs, or unusable verification logic.

\textbf{Independent Solver evaluation.} Each self-validated task is attempted once by DeepSeek-V4-Pro as the strong Solver and once by Qwen3.5-35B-A3B as the weak Solver. To elicit the target behavior, we provide both Solvers with the same explicit prompt. Before each skill-guided execution stage, the prompt instructs each Solver to read the supplied skills and express a task-specific plan in the Think with Skills pseudocode format, specifying concrete operations, inputs and outputs, control-flow branches, intermediate artifacts, dependencies, and verification conditions. Each Solver then executes the task with this pseudocode retained in its context, producing an aligned plan-and-execution trajectory rather than a post-hoc rationale. The two Solvers start from separate copies of the same candidate state, and the task-specific verification script evaluates their outputs independently.

\textbf{Outcome routing and task revision.} A task is accepted when the strong Solver succeeds and the weak Solver fails, indicating that it is solvable but remains informative for training the target model. If both Solvers succeed, an Improver Agent raises the difficulty. If both fail, a Difficulty Reducer Agent repairs the task or lowers its difficulty. If the strong Solver fails while the weak Solver succeeds, the inconsistent candidate is discarded. The Improver and Difficulty Reducer may revise the task description, input files, and verification script. Each revised task must pass self-validation again before both Solvers retry it. A candidate undergoes at most three revisions before acceptance or rejection, as detailed in Section~\ref{sec:4.1}.

For each accepted task, we retain the sampled skills, task description, input files, verification script, and successful strong-Solver trajectory. The trajectory records the pseudocode produced before each skill-guided stage together with the subsequent execution process, and is used as supervision in the first training stage. Weak-Solver trajectories are used only for filtering and are not included in the training corpus. During GRPO training, the retained verification script supplies an executable task-level reward.

\subsection{Two-Stage Training}
\label{sec:3.3}

\textbf{Supervised fine-tuning.} We first fine-tune the target model on the collected trajectories. For each example, the task instruction, input files, relevant skills, and available tools form the input context. Each training example includes a target trajectory in which a task-specific pseudocode plan appears immediately before the corresponding skill-guided execution stage. This ordering directly supervises the behavior required by Think with Skills. The model learns to convert each provided skill into a concrete procedure before using it to guide tool interactions.

\textbf{Group Relative Policy Optimization.} Starting from the supervised checkpoint, we further train the model with standard outcome-supervised GRPO \citep{shao2024deepseekmath}. For each task, the policy samples a group of complete agent rollouts in which the model generates task-specific pseudocode and then interacts with the environment. The task-specific verifier assigns a reward to the final outcome of each rollout, and GRPO uses group-normalized rewards as relative advantages under its standard clipped objective with KL regularization. Because the verifier scores the final task outcome, GRPO propagates the resulting reward through the entire rollout, updating the policy for both pseudocode generation and subsequent task execution. The single-attempt strong and weak Solver evaluations in Section~\ref{sec:3.2} are used only for data construction. GRPO samples a separate group of rollouts during training. Training hyperparameters are reported in Section~\ref{sec:4.2}.

\section{Experiments}
\label{sec:4}

\subsection{Constructed Training Corpus}
\label{sec:4.1}

\subsubsection*{Skill Source and Task Construction}

We construct the training corpus from connected skill sequences sampled from Graph-of-Skills after excluding skills reserved for SkillsBench evaluation. Of the 2,000 skills in the original graph, we exclude the 174 that appear in SkillsBench. Graph-of-Skills connects skills through dependency, workflow, semantic, and alternative relations \citep{liu2026graphskills}. From the remaining graph, we sample connected sequences of one to three skills and use each sequence to condition the generation of a task that uses the selected skills.

Following the pipeline in Section~\ref{sec:3.2}, the Draft Agent, implemented with DeepSeek-V4-Pro, generates each task, its input files, and its verifier. DeepSeek-V4-Pro and Qwen3.5-35B-A3B serve as the strong and weak Solvers and attempt each candidate once. The Improver and Difficulty Reducer use DeepSeek-V4-Pro, and a candidate undergoes at most 3 revisions before acceptance or rejection.

\subsection{Experimental Setup}
\label{sec:4.2}

We evaluate on SkillsBench using its official tasks, expert-written skills, environments, and task-success criteria. The primary metric is task success rate. Think with Skills is initialized from Qwen3.5-35B-A3B and trained with SFT followed by GRPO. Our evaluations use Claude Code as the inference-time agent framework. Table~\ref{tab:1} reports the agent framework, model, and SkillsBench result for each configuration. External results may use different inference frameworks and evaluation settings.

\subsection{Main Results}
\label{sec:4.3}

Table~\ref{tab:1} presents Think with Skills and its unmodified Qwen3.5-35B-A3B base model first, followed by selected reference configurations. Officially reported scores are marked with \textdagger{} and retain their original inference frameworks. All five of our evaluation rows use Claude Code.

\begin{table*}[!t]
\centering
\caption{SkillsBench results with expert skills.}
\label{tab:1}
\small
\renewcommand{\arraystretch}{1.2}
\begin{tabularx}{\linewidth}{@{}>{\raggedright\arraybackslash}p{0.24\linewidth} >{\raggedright\arraybackslash}X >{\raggedleft\arraybackslash}p{0.22\linewidth}@{}}
\toprule
\textbf{Agent framework} & \textbf{Model} & \textbf{SkillsBench result (\%)} \\
\midrule
\multirow{5}{*}{Claude Code} & \textbf{Think with Skills (Qwen3.5-35B-A3B)} & \textbf{62.0 (+27.0 pp)} \\
\cmidrule(lr){2-3}
& Qwen3.5-35B-A3B & 35.0 \\
\cmidrule(lr){2-3}
& Kimi K2.5 & 62.5 \\
\cmidrule(lr){2-3}
& DeepSeek V4 Pro & 61.67 \\
\cmidrule(lr){2-3}
& MiniMax M2.7 & 42.76 \\
\midrule
\multirow{2}{*}{OpenHands} & GLM 5.1 & 58.4\textdagger{} \\
\cmidrule(lr){2-3}
& Opus 4.8 & 54.1\textdagger{} \\
\midrule
Gemini CLI & Gemini 3 Flash & 54.6\textdagger{} \\
\bottomrule
\end{tabularx}

\begin{minipage}{\linewidth}
\footnotesize \textit{Note.} \textdagger{} denotes an \href{https://huggingface.co/datasets/benchflow/skillsbench-leaderboard/blob/main/leaderboard/skillsbench/v1.1/official.csv}{official SkillsBench v1.1 result} on 87 tasks. External scores are included for reference because agent frameworks and evaluation settings may differ.
\end{minipage}
\end{table*}

Our evaluations of MiniMax M2.7, DeepSeek V4 Pro and Kimi K2.5 use Claude Code, with success rates of 42.76\%, 61.67\% and 62.5\%, respectively. These scores retain the precision supplied in our evaluation records.

Think with Skills increases the success rate recorded for Qwen3.5-35B-A3B from 35.0\% to 62.0\%, an absolute gain of 27.0 percentage points. This controlled comparison against the base model is the primary evidence for the method's improvement. External scores are included as reference values and do not establish a controlled ranking across different agent frameworks and evaluation settings.

\section{Analysis}
\label{sec:5}

The main results evaluate Think with Skills as a complete system. We next isolate the effects of the pseudocode representation and the two post-training stages. We then examine how the number of skills per training task affects downstream performance. All variants are initialized from the same Qwen3.5-35B-A3B base model and use the same SkillsBench evaluation tasks, tools, and interaction budget unless otherwise stated.

\subsection{Effects of Pseudocode Planning and Post-Training}
\label{sec:5.1}

Table~\ref{tab:2} compares four configurations of Qwen3.5-35B-A3B. The configurations include direct skill use by the base model, pseudocode prompting applied to the base model without additional training, SFT, and SFT followed by GRPO. The direct and prompted base-model settings receive the same task inputs, tools, and expert-written skills under the same execution budget. In the prompt-only configuration, an explicit prompt instructs the base model to express its plan in the Think with Skills pseudocode format before execution, without any parameter updates. SFT jointly supervises pseudocode plans and their corresponding execution trajectories, as described in Section~\ref{sec:3.3}. Both post-trained configurations generate pseudocode before execution.

\begin{table}[!htbp]
\centering
\caption{SkillsBench task success rates for Qwen3.5-35B-A3B under direct skill use, pseudocode prompting, SFT, and SFT followed by GRPO.}
\label{tab:2}
\small
\renewcommand{\arraystretch}{1.2}
\begin{tabularx}{\linewidth}{@{}>{\raggedright\arraybackslash}X r@{}}
\toprule
\textbf{Model / Training Configuration} & \shortstack{\textbf{SkillsBench}\\\textbf{Score (\%)}} \\
\midrule
Qwen3.5-35B-A3B (base model, direct skill use) & 35.0 \\
Qwen3.5-35B-A3B + pseudocode prompting & 45.0 \\
Qwen3.5-35B-A3B + SFT & 60.0 \\
Qwen3.5-35B-A3B + SFT + GRPO (Think with Skills) & \textbf{62.0} \\
\bottomrule
\end{tabularx}
\end{table}

Pseudocode prompting raises the base model's success rate from 35.0\% to 45.0\%, a gain of 10.0 percentage points without additional training. SFT raises success to 60.0\%, adding 15.0 percentage points over prompting alone. Continuing from the SFT checkpoint with GRPO raises success to 62.0\%, a further gain of 2.0 percentage points. The complete Think with Skills procedure therefore improves success by 27.0 percentage points over the base model.

\subsection{Effect of the Number of Skills per Training Task}
\label{sec:5.2}

We compare training on tasks involving a single skill with training on tasks involving multiple skills. The single-skill setting uses 1,000 training examples, each involving one skill. The multi-skill setting uses 900 training examples, each involving more than one skill. Table~\ref{tab:3} reports downstream SkillsBench task success for the two training-data settings using Qwen3.5-35B-A3B as the base model.

\begin{table}[!htbp]
\centering
\caption{Comparison of single-skill and multi-skill training on SkillsBench.}
\label{tab:3}
\small
\renewcommand{\arraystretch}{1.2}
\begin{tabularx}{\linewidth}{@{}>{\raggedright\arraybackslash}X r r@{}}
\toprule
\textbf{Skills per Training Task} & \shortstack{\textbf{Training}\\\textbf{Examples}} & \shortstack{\textbf{SkillsBench}\\\textbf{Success (\%)}} \\
\midrule
Single skill (1) & 901 & 55.0 \\
Multiple skills (>1) & 913 & \textbf{60.0} \\
\bottomrule
\end{tabularx}
\end{table}

Training on multi-skill tasks yields a success rate of 60.0\%, compared with 55.0\% for single-skill tasks, an absolute gain of 5.0 percentage points. The multi-skill setting performs better despite using 100 fewer training examples. This result suggests that training on tasks requiring coordination across skills can improve downstream skill use.

\section{Conclusion and Limitations}
\label{sec:6}

We introduced Think with Skills, a training framework that teaches language model agents to convert expert-written skills into task-specific pseudocode before execution. The method combines a verifiable, difficulty-calibrated data construction pipeline with supervised fine-tuning and outcome-based reinforcement learning. On SkillsBench, it raises the success rate of Qwen3.5-35B-A3B from 35.0\% to 62.0\%, showing that explicit proceduralization can make expert guidance substantially more useful to an agent.

\bibliographystyle{icml2025}
\bibliography{references}

@article{anderson1982acquisition,
  author  = {John R. Anderson},
  title   = {Acquisition of Cognitive Skill},
  journal = {Psychological Review},
  year    = {1982},
  volume  = {89},
  number  = {4},
  pages   = {369--406},
  doi     = {10.1037/0033-295X.89.4.369}
}

@inproceedings{yao2023react,
  author    = {Shunyu Yao and Jeffrey Zhao and Dian Yu and Nan Du and
               Izhak Shafran and Karthik Narasimhan and Yuan Cao},
  title     = {{ReAct}: Synergizing Reasoning and Acting in Language Models},
  booktitle = {The Eleventh International Conference on Learning Representations},
  year      = {2023},
  url       = {https://openreview.net/forum?id=WE_vluYUL-X}
}

@misc{li2026skillsbench,
  author       = {Xiangyi Li and Wenbo Chen and Yimin Liu and Shenghan Zheng and
                  Xiaokun Chen and Yifeng He and Yubo Li and Bingran You and
                  Haotian Shen and Jiankai Sun and Shuyi Wang and Qunhong Zeng and
                  Di Wang and Xuandong Zhao and Yuanli Wang and Roey Ben Chaim and
                  Zonglin Di and Yipeng Gao and Junwei He and Yizhuo He and
                  Liqiang Jing and Luyang Kong and Xin Lan and Jiachen Li and
                  Songlin Li and Yijiang Li and Yueqian Lin and Xinyi Liu and
                  Xuanqing Liu and Haoran Lyu and Ze Ma and Bowei Wang and
                  Runhui Wang and Tianyu Wang and Wengao Ye and Yue Zhang and
                  Hanwen Xing and Yiqi Xue and Steven Dillmann and Han-chung Lee},
  title        = {{SkillsBench}: Benchmarking How Well Agent Skills Work Across Diverse Tasks},
  year         = {2026},
  eprint       = {2602.12670},
  archiveprefix = {arXiv},
  primaryclass = {cs.AI},
  note         = {Version 1},
  url          = {https://arxiv.org/abs/2602.12670v1}
}

@inproceedings{zheng2024executing,
  author    = {Xin Zheng and Qiming Zhu and Hongyu Lin and Yaojie Lu and
               Xianpei Han and Le Sun},
  title     = {Executing Natural Language-Described Algorithms with Large Language Models:
               An Investigation},
  booktitle = {Proceedings of the 2024 Joint International Conference on
               Computational Linguistics, Language Resources and Evaluation
               ({LREC-COLING} 2024)},
  year      = {2024},
  month     = may,
  pages     = {6752--6837},
  address   = {Torino, Italia},
  publisher = {ELRA and ICCL},
  url       = {https://aclanthology.org/2024.lrec-main.596/}
}

@inproceedings{wei2022chain,
  author    = {Jason Wei and Xuezhi Wang and Dale Schuurmans and Maarten Bosma and
               Brian Ichter and Fei Xia and Ed H. Chi and Quoc V. Le and Denny Zhou},
  title     = {Chain-of-Thought Prompting Elicits Reasoning in Large Language Models},
  booktitle = {Advances in Neural Information Processing Systems},
  year      = {2022},
  volume    = {35},
  pages     = {24824--24837},
  publisher = {Curran Associates, Inc.},
  doi       = {10.52202/068431-1800},
  url       = {https://proceedings.neurips.cc/paper/2022/hash/9d5609613524ecf4f15af0f7b31abca4-Abstract.html}
}

@inproceedings{zhou2023least,
  author    = {Denny Zhou and Nathanael Sch{\"a}rli and Le Hou and Jason Wei and
               Nathan Scales and Xuezhi Wang and Dale Schuurmans and Claire Cui and
               Olivier Bousquet and Quoc V. Le and Ed H. Chi},
  title     = {Least-to-Most Prompting Enables Complex Reasoning in Large Language Models},
  booktitle = {The Eleventh International Conference on Learning Representations},
  year      = {2023},
  url       = {https://openreview.net/forum?id=WZH7099tgfM}
}

@inproceedings{wang2023plan,
  author    = {Lei Wang and Wanyu Xu and Yihuai Lan and Zhiqiang Hu and
               Yunshi Lan and Roy Ka-Wei Lee and Ee-Peng Lim},
  title     = {Plan-and-Solve Prompting: Improving Zero-Shot Chain-of-Thought
               Reasoning by Large Language Models},
  booktitle = {Proceedings of the 61st Annual Meeting of the Association for
               Computational Linguistics (Volume 1: Long Papers)},
  year      = {2023},
  month     = jul,
  pages     = {2609--2634},
  address   = {Toronto, Canada},
  publisher = {Association for Computational Linguistics},
  doi       = {10.18653/v1/2023.acl-long.147},
  url       = {https://aclanthology.org/2023.acl-long.147/}
}

@inproceedings{gao2023pal,
  author    = {Luyu Gao and Aman Madaan and Shuyan Zhou and Uri Alon and
               Pengfei Liu and Yiming Yang and Jamie Callan and Graham Neubig},
  title     = {{PAL}: Program-Aided Language Models},
  booktitle = {Proceedings of the 40th International Conference on Machine Learning},
  year      = {2023},
  editor    = {Andreas Krause and Emma Brunskill and Kyunghyun Cho and
               Barbara Engelhardt and Sivan Sabato and Jonathan Scarlett},
  volume    = {202},
  series    = {Proceedings of Machine Learning Research},
  pages     = {10764--10799},
  publisher = {PMLR},
  url       = {https://proceedings.mlr.press/v202/gao23f.html}
}

@article{chen2023program,
  author  = {Wenhu Chen and Xueguang Ma and Xinyi Wang and William W. Cohen},
  title   = {Program of Thoughts Prompting: Disentangling Computation from
             Reasoning for Numerical Reasoning Tasks},
  journal = {Transactions on Machine Learning Research},
  year    = {2023},
  issn    = {2835-8856},
  url     = {https://openreview.net/forum?id=YfZ4ZPt8zd}
}

@inproceedings{chae2024language,
  author    = {Hyungjoo Chae and Yeonghyeon Kim and Seungone Kim and
               Kai Tzu-iunn Ong and Beong-woo Kwak and Moohyeon Kim and
               Sunghwan Kim and Taeyoon Kwon and Jiwan Chung and Youngjae Yu and
               Jinyoung Yeo},
  title     = {Language Models as Compilers: Simulating Pseudocode Execution
               Improves Algorithmic Reasoning in Language Models},
  booktitle = {Proceedings of the 2024 Conference on Empirical Methods in
               Natural Language Processing},
  year      = {2024},
  month     = nov,
  pages     = {22471--22502},
  address   = {Miami, Florida, USA},
  publisher = {Association for Computational Linguistics},
  doi       = {10.18653/v1/2024.emnlp-main.1253},
  url       = {https://aclanthology.org/2024.emnlp-main.1253/}
}

@inproceedings{zelikman2022star,
  author    = {Eric Zelikman and Yuhuai Wu and Jesse Mu and Noah Goodman},
  title     = {{STaR}: Bootstrapping Reasoning With Reasoning},
  booktitle = {Advances in Neural Information Processing Systems},
  year      = {2022},
  volume    = {35},
  pages     = {15476--15488},
  publisher = {Curran Associates, Inc.},
  doi       = {10.52202/068431-1126},
  url       = {https://proceedings.neurips.cc/paper_files/paper/2022/hash/639a9a172c044fbb64175b5fad42e9a5-Abstract-Conference.html}
}

@inproceedings{hsieh2023distilling,
  author    = {Cheng-Yu Hsieh and Chun-Liang Li and Chih-kuan Yeh and
               Hootan Nakhost and Yasuhisa Fujii and Alex Ratner and
               Ranjay Krishna and Chen-Yu Lee and Tomas Pfister},
  title     = {Distilling Step-by-Step! Outperforming Larger Language Models
               with Less Training Data and Smaller Model Sizes},
  booktitle = {Findings of the Association for Computational Linguistics:
               {ACL} 2023},
  year      = {2023},
  month     = jul,
  pages     = {8003--8017},
  address   = {Toronto, Canada},
  publisher = {Association for Computational Linguistics},
  doi       = {10.18653/v1/2023.findings-acl.507},
  url       = {https://aclanthology.org/2023.findings-acl.507/}
}

@inproceedings{liu2024skillact,
  author    = {Anthony Zhe Liu and Jongwook Choi and Sungryull Sohn and
               Yao Fu and Jaekyeom Kim and Dong-Ki Kim and Xinhe Wang and
               Jaewon Yoo and Honglak Lee},
  title     = {{SkillAct}: Using Skill Abstractions Improves {LLM} Agents},
  booktitle = {ICML 2024 Workshop on LLMs and Cognition},
  year      = {2024},
  url       = {https://openreview.net/forum?id=6LG3cIRrF4}
}

@article{wang2024voyager,
  author  = {Guanzhi Wang and Yuqi Xie and Yunfan Jiang and Ajay Mandlekar and
             Chaowei Xiao and Yuke Zhu and Linxi Fan and Anima Anandkumar},
  title   = {{VOYAGER}: An Open-Ended Embodied Agent with Large Language Models},
  journal = {Transactions on Machine Learning Research},
  year    = {2024},
  issn    = {2835-8856},
  url     = {https://openreview.net/forum?id=ehfRiF0R3a}
}

@inproceedings{zhao2024expel,
  author    = {Andrew Zhao and Daniel Huang and Quentin Xu and Matthieu Lin and
               Yong-Jin Liu and Gao Huang},
  title     = {{ExpeL}: {LLM} Agents Are Experiential Learners},
  booktitle = {Proceedings of the AAAI Conference on Artificial Intelligence},
  year      = {2024},
  volume    = {38},
  number    = {17},
  pages     = {19632--19642},
  doi       = {10.1609/aaai.v38i17.29936},
  url       = {https://ojs.aaai.org/index.php/AAAI/article/view/29936}
}

@misc{wang2025sage,
  author        = {Jiongxiao Wang and Qiaojing Yan and Yawei Wang and Yijun Tian and
                   Soumya Smruti Mishra and Zhichao Xu and Megha Gandhi and
                   Panpan Xu and Lin Lee Cheong},
  title         = {Reinforcement Learning for Self-Improving Agent with Skill Library},
  year          = {2025},
  eprint        = {2512.17102},
  archiveprefix = {arXiv},
  primaryclass  = {cs.AI},
  doi           = {10.48550/arXiv.2512.17102},
  url           = {https://arxiv.org/abs/2512.17102}
}

@misc{pan2026anything2skill,
  author        = {Qianjun Pan and Yutao Yang and Junsong Li and Jie Zhou and
                   Kai Chen and Xin Li and Qin Chen and Liang He},
  title         = {{Anything2Skill}: Compiling External Knowledge into Reusable Skills for Agents},
  year          = {2026},
  eprint        = {2606.09316},
  archiveprefix = {arXiv},
  primaryclass  = {cs.AI},
  doi           = {10.48550/arXiv.2606.09316},
  url           = {https://arxiv.org/abs/2606.09316}
}

@misc{su2026sra,
  author        = {Weihang Su and Jianming Long and Qingyao Ai and Yichen Tang and
                   Changyue Wang and Yiteng Tu and Yiqun Liu},
  title         = {Skill Retrieval Augmentation for Agentic {AI}},
  year          = {2026},
  eprint        = {2604.24594},
  archiveprefix = {arXiv},
  primaryclass  = {cs.AI},
  doi           = {10.48550/arXiv.2604.24594},
  url           = {https://arxiv.org/abs/2604.24594}
}

@misc{liu2026skillrevise,
  author        = {Yuxuan Liu and Zhaochen Su and Lingyun Xie and Yuhao Zhang and
                   Qing Zong and Jiahe Guo and Zhongwei Xie and Yiyan Ji and
                   Yauwai Yim and Hongyu Luo and Xiyu Ren and Ruan Chenyu and
                   Haoran Li and Yangqiu Song},
  title         = {{SkillRevise}: Improving {LLM}-Authored Agent Skills via
                   Trace-Conditioned Skill Revision},
  year          = {2026},
  eprint        = {2606.01139},
  archiveprefix = {arXiv},
  primaryclass  = {cs.AI},
  doi           = {10.48550/arXiv.2606.01139},
  note          = {Accepted to Findings of EMNLP 2026},
  url           = {https://arxiv.org/abs/2606.01139}
}

@misc{zhao2026skillcomposer,
  author        = {Xinyu Zhao and Zhen Tan and Vaishnav Tadiparthi and
                   Nakul Agarwal and Kwonjoon Lee and Ehsan Moradi Pari and
                   Hossein Nourkhiz Mahjoub and Tianlong Chen},
  title         = {Generative Skill Composition for {LLM} Agents},
  year          = {2026},
  eprint        = {2606.32025},
  archiveprefix = {arXiv},
  primaryclass  = {cs.CL},
  doi           = {10.48550/arXiv.2606.32025},
  url           = {https://arxiv.org/abs/2606.32025}
}

@misc{shen2026grounded,
  author        = {Jianghan Shen and Zhenjie Liu and Yue Li and Jie Huang and
                   Siqi Luo and Yiming Cheng and Yizhi Yao and Kaijie Zhang and
                   Cheng Tang and Minghui Zhang and Ming Hu and Yirong Chen and
                   Ziyan Huang},
  title         = {Beyond Instruction Following: Learning Grounded
                   Skill-Following with Skill Contracts},
  year          = {2026},
  eprint        = {2610.05161},
  archiveprefix = {arXiv},
  primaryclass  = {cs.AI},
  url           = {https://arxiv.org/abs/2610.05161}
}

@inproceedings{liu2024rafa,
  author    = {Zhihan Liu and Hao Hu and Shenao Zhang and Hongyi Guo and
               Shuqi Ke and Boyi Liu and Zhaoran Wang},
  title     = {Reason for Future, Act for Now: A Principled Architecture for
               Autonomous {LLM} Agents},
  booktitle = {Proceedings of the 41st International Conference on Machine Learning},
  year      = {2024},
  editor    = {Ruslan Salakhutdinov and Zico Kolter and Katherine Heller and
               Adrian Weller and Nuria Oliver and Jonathan Scarlett and
               Felix Berkenkamp},
  volume    = {235},
  series    = {Proceedings of Machine Learning Research},
  pages     = {31186--31261},
  publisher = {PMLR},
  url       = {https://proceedings.mlr.press/v235/liu24ab.html}
}

@inproceedings{erdogan2025planandact,
  author    = {Lutfi Eren Erdogan and Nicholas Lee and Sehoon Kim and
               Suhong Moon and Hiroki Furuta and Gopala Anumanchipalli and
               Kurt Keutzer and Amir Gholami},
  title     = {Plan-and-Act: Improving Planning of Agents for Long-Horizon Tasks},
  booktitle = {Proceedings of the 42nd International Conference on Machine Learning},
  year      = {2025},
  editor    = {Aarti Singh and Maryam Fazel and Daniel Hsu and
               Simon Lacoste-Julien and Felix Berkenkamp and Tegan Maharaj and
               Kiri Wagstaff and Jerry Zhu},
  volume    = {267},
  series    = {Proceedings of Machine Learning Research},
  pages     = {15419--15462},
  publisher = {PMLR},
  url       = {https://proceedings.mlr.press/v267/erdogan25a.html}
}

@inproceedings{parmar2025plantuning,
  author    = {Mihir Parmar and Palash Goyal and Xin Liu and Yiwen Song and
               Mingyang Ling and Chitta Baral and Hamid Palangi and Tomas Pfister},
  title     = {{PLAN-TUNING}: Post-Training Language Models to Learn
               Step-by-Step Planning for Complex Problem Solving},
  booktitle = {Proceedings of the 2025 Conference on Empirical Methods in
               Natural Language Processing},
  year      = {2025},
  month     = nov,
  pages     = {21419--21433},
  address   = {Suzhou, China},
  publisher = {Association for Computational Linguistics},
  doi       = {10.18653/v1/2025.emnlp-main.1087},
  url       = {https://aclanthology.org/2025.emnlp-main.1087/}
}

@inproceedings{liu2026cope,
  author    = {Huanxi Liu and Kun Hu and Qiang Wang and Yuanzhao Zhai and
               Feng Dawei and Bo Ding and Huaimin Wang},
  title     = {{CoPE}: A Framework for Optimizing Coordination between Planning
               and Execution in {LLM}-Based Agents},
  booktitle = {Proceedings of the 43rd International Conference on Machine Learning},
  year      = {2026},
  editor    = {Tong Zhang and Miroslav Dudik and Martin Jaggi and Alekh Agarwal and
               Sharon Li and Dale Schuurmans and Jerry Zhu and Felix Berkenkamp and
               Hanze Dong and Alberto Bietti},
  volume    = {306},
  series    = {Proceedings of Machine Learning Research},
  pages     = {76571--76601},
  publisher = {PMLR},
  url       = {https://proceedings.mlr.press/v306/liu26az.html}
}

@article{liu2026graphskills,
  author  = {Dawei Liu and Zongxia Li and Hongyang Du and Xiyang Wu and Shihang Gui and Yongbei Kuang and Lichao Sun},
  title   = {Graph-of-Skills: Dependency-Aware Structural Retrieval for Massive Agent Skills},
  journal = {arXiv preprint arXiv:2604.05333},
  year    = {2026},
  url     = {https://arxiv.org/abs/2604.05333}
}

@article{shao2024deepseekmath,
  author  = {Zhihong Shao and Peiyi Wang and Qihao Zhu and Runxin Xu and Junxiao Song and Xiao Bi and Haowei Zhang and Mingchuan Zhang and Y. K. Li and Y. Wu and Daya Guo},
  title   = {DeepSeekMath: Pushing the Limits of Mathematical Reasoning in Open Language Models},
  journal = {arXiv preprint arXiv:2402.03300},
  year    = {2024},
  url     = {https://arxiv.org/abs/2402.03300}
}
\end{document}